\pdfoutput=1

\documentclass[11pt]{article}

\usepackage[preprint]{acl}

\usepackage{array}

\usepackage{times}
\usepackage{latexsym}

\usepackage[T1]{fontenc}

\usepackage[utf8]{inputenc}

\usepackage{microtype}

\usepackage{inconsolata}

\usepackage{graphicx}

\usepackage{tikz}        %
\usetikzlibrary{arrows}  %
\usetikzlibrary{shapes}  %
\usetikzlibrary{positioning}  %
\usepackage{pgffor}      %
\usetikzlibrary{calc}  %

\usepackage{enumitem}    %
\usepackage[font=small]{caption}  %
\usepackage[normalem]{ulem}

\usepackage{microtype}   %
\usepackage{xcolor}      %

\usepackage{multirow}

\usepackage{listings}
\usepackage[breakable]{tcolorbox}
\usepackage{booktabs}

\title{All That Glitters is Not Novel: Plagiarism in AI Generated Research}

\author{Tarun Gupta \\
  Indian Institute of Science \\
  Bengaluru, KA, India \\
  \texttt{tarungupta@iisc.ac.in} \\\And
  Danish Pruthi \\
  Indian Institute of Science \\
  Bengaluru, KA, India \\
  \texttt{danishp@iisc.ac.in} \\}

\begin{document}
\maketitle

\begin{abstract}
    Automating scientific research is considered the final frontier of science. Recently, several papers claim autonomous research agents can generate novel research ideas. Amidst the prevailing optimism, we document a critical concern: a considerable fraction of such research documents are smartly plagiarized. Unlike past efforts where experts evaluate the novelty and feasibility of research ideas, we request $13$ experts to operate under a different situational logic: to identify similarities between LLM-generated research documents and existing work. Concerningly, the experts identify $24\%$ of the $50$ evaluated research documents to be either paraphrased (with one-to-one methodological mapping), or significantly borrowed from existing work. 
    These reported instances are cross-verified by authors of the source papers.
    The remaining $76\%$ of documents show varying degrees of similarity with existing work, with only a small fraction appearing completely novel.
    Problematically, these LLM-generated research documents do not acknowledge original sources, and bypass inbuilt plagiarism detectors. Lastly, through controlled experiments we show that automated plagiarism detectors are inadequate at catching plagiarized ideas from such systems. We recommend a careful assessment of LLM-generated research, and discuss the implications of our findings on academic publishing.\footnote{Our code, along with  expert-provided scores and explanations for each proposal, is available at: \url{https://github.com/tarun360/AI-Papers-Plagiarism/}}
\end{abstract}

\section{Introduction}
\label{sec:intro}

\begin{figure*}[t]
    \centering
    \begin{tikzpicture}[
        box/.style={draw, rounded corners=15pt, minimum width=2cm, minimum height=1.25cm, 
            align=center, font=\scriptsize, inner sep=5pt},
        violetbox/.style={box, fill=violet!10, draw=violet!30},
        rightbox/.style={box, fill=orange!5, draw=orange!20},
        redbox/.style={box, fill=red!15, draw=red!50},
        arrow/.style={->, >=latex, thick, draw=gray!40, shorten >=2pt, shorten <=2pt}
    ]
        \begin{scope}[scale=0.8]
        
        \node[violetbox] (b) at (0,2) {
        $12$ NLP topics (Table~\ref{tab:research-topics})\\$\times$ $3$ = $36$ research proposals\\ generated from \citet{si2024can} \\
        + \\
        $4$ research proposals showcased\\in \citet{si2024can} \\
        + \\
        $10$ research papers showcased in\\ \citet{lu2024ai}   
        };
        
        \node[violetbox] (o1) at (5,2) {Total $50$\\ research documents};
        \node[rightbox] (o2) at (9.75,2) {Expert assessment: identify\\source papers and assign score\\ ($1$-$5$) (scores defined in Table~\ref{tab:similarity-scores})};
        \node[redbox] (o3) at (15.25,2) {Verification by source\\paper's author.\\24.0\% verified plagiarism\\ (similarity score $4+$)};

        \node[inner sep=0] (search) at (7,4) {\includegraphics[width=1cm]{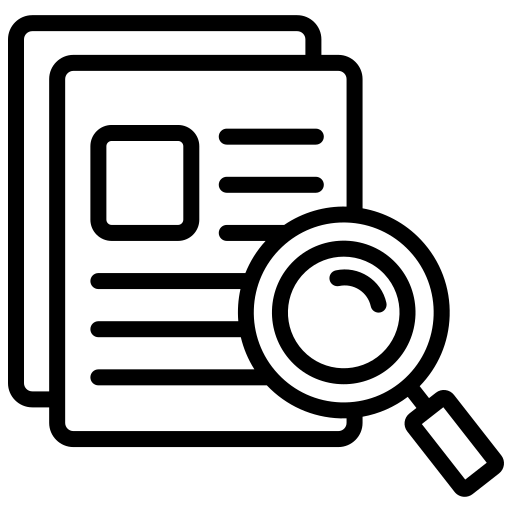}};
        
        \node[text width=3cm, font=\tiny, align=center] (plag-text) at (7,3) {$13$ experts looking\\for plagiarism};
        
        \draw[arrow] (b.east) -- (o1.west);
        
        \draw[arrow] (o1.east) -- (o2.west);
        \draw[arrow] (o2.east) -- (o3.west);
        
        \end{scope}
    \end{tikzpicture}
    \caption{Overview of our expert-led evaluation for detecting plagiarism in LLM-generated research proposals. Unlike prior work, participants in our study are instructed to actively search for potential sources of plagiarism.}
    \label{fig:expert-evaluation-setup}
\end{figure*}
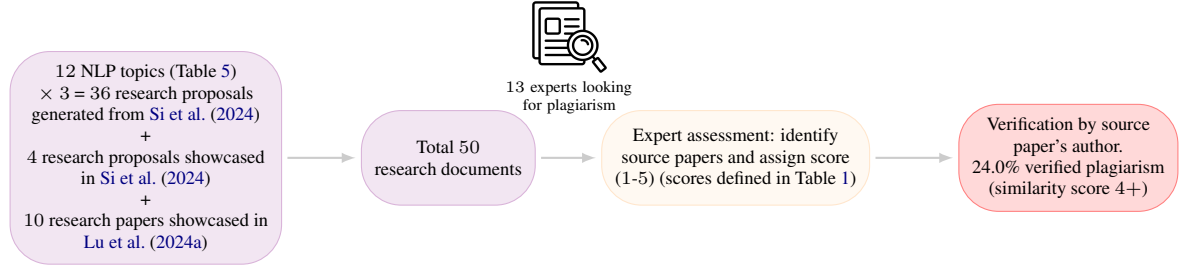

Automating research and discovering new knowledge 
has been a longstanding aspiration. 
The first step of 
scientific research is 
coming up with a bold hypothesis or a conjecture \citep{popper2014conjectures}. 
Automating this step is a crucial component of automating scientific research.
Recent research presents a positive case of
LLMs' ability to 
generate novel scientific contributions---be they hypotheses, or proposals or papers \citep{li2024chain,lu2024ai, baek2024researchagent, li2024mlr, wang2023scimon, yang2023large, li2024learning, weng2024cycleresearcher}. 

Understandably, 
evaluating the novelty LLM-generated ideas 
is challenging, especially 
given the subjective nature of scientific innovation. 
Previous studies 
evaluate novelty either through automated LLM-based judges \citep{lu2024ai}, or rely on small set of 
experts \citep{li2024chain, baek2024researchagent, li2024mlr, wang2023scimon, yang2023large, li2024learning, weng2024cycleresearcher}. 
Notably, the most 
rigorous evaluation to date 
engaged  experts %
to evaluate $81$ LLM-generated research proposals, 
implementing strict controls for confounding factors \citep{si2024can}. 
Their study leads to an  
important finding: 
human experts find LLM-generated research 
proposals 
to be \emph{more novel} than human-written ones.
The study 
also publicly releases 
four 
exemplar LLM-generated proposals,
holding back others to be used for future work.

In this work, we conduct an expert-led 
evaluation 
where participants 
are instead  
instructed to presume plagiarism and 
actively search for it in LLM-generated research documents. 
This situational logic \citep{popper2013poverty, hoover2016situational} contrasts 
with prior assessment of LLM generated ideas,
where experts evaluate a shuffled set 
of LLM and human-generated documents, 
presuming no deliberate plagiarism 
and scoring them on novelty and other factors, such as excitement and feasibility. 
In total, 
experts in our study evaluate 
$50$ LLM-generated research documents, 
including $10$ 
exemplar papers 
generated by the ``The AI Scientist''~\citep{lu2024ai}, 
four 
public research proposals from \citet{si2024can}, 
and $36$ fresh proposals.

We request 
the experts in our study 
to identify a topic of their expertise, 
and 
share with
them $5$ 
research proposals 
related to that topic. 
We 
generate these proposals through 
the code made available by \citet{si2024can}. 
The experts are then requested 
to score any $3$ of the 
$5$ research proposals 
on a  
scale of $1$-$5$, 
where $5$ indicates 
direct 
copying---that is, there exists 
a one-to-one mapping 
between the LLM proposed methodology
and existing methods 
in one or two closely related prior papers.
A score of $4$ 
denotes that 
a significant portion of the LLM proposed method
was borrowed from $2$-$3$ prior works without credit.
On the other extreme, 
a score of $1$ is reserved 
for cases when 
experts 
find the proposal to be completely original
(see Table~\ref{tab:similarity-scores}
for the rubric).

Our expert-led analysis reveals that $14.0\%$ of LLM-generated research documents 
receive a score of $5$ and another $10.0\%$ 
documents receive a score of $4$. 
We consider both scores $4$ and $5$ as instances of plagiarism, 
totaling $24.0\%$ of proposals with noticeable plagiarism. 
These scores are verified by emailing the authors of referenced papers. 
Notably, several
previously 
showcased exemplars of 
LLM generated research \citep{si2024can,lu2024ai,yamada2025ai} are either found to be plagiarized or substantially similar to existing work.
Our setup 
and key findings 
are illustrated in Figure~\ref{fig:expert-evaluation-setup}.

It is important to note 
that the examined documents
do not 
acknowledge the original sources
and the high degree of 
similarity with past 
work 
is not caught by 
inbuilt plagiarism detection systems.
Typically,  
these inbuilt plagiarism detectors 
use large language models,  
which have access 
to the Semantic Scholar API 
to retrieve similar papers
\citep{si2024can, lu2024ai, li2024chain}.\footnote{We refer to this approach as Semantic Scholar Augmented Generation (SSAG) throughout the paper.}
To systematically evaluate automated detection methods, 
we create a synthetic dataset of 
research proposals intentionally plagiarized from existing papers.
Using this controlled dataset, 
we evaluate common automated detection methods, 
including SSAG, OpenScholar (embedding-based search) \citep{asai2024openscholar}, and a
commercial service \citep{turnitin}, and find them inadequate for detecting plagiarism in LLM-generated research proposals.

\begin{table*}[t]
    \centering
    \small
    \begin{tabular}{@{}cl@{}}
        \toprule
        \textbf{Score} & \textbf{Description} \\
        \midrule \vspace{1.5mm}
        5 & \begin{tabular}[c]{@{}l@{}}\textbf{Direct Copy:} One-to-one mapping between the LLM proposed methodology \\ and existing methods in one or two closely related prior papers. \end{tabular} \\ \vspace{1.5mm}
        4 & \begin{tabular}[c]{@{}l@{}}\textbf{Combined Borrowing:} A significant portion of LLM proposed method is a \\ mix-and-match  from two-to-three prior works. \end{tabular} \\ \vspace{1.5mm}
        3 & \begin{tabular}[c]{@{}l@{}}\textbf{Partial Overlap:} The LLM proposed method bears decent similarity with some existing \\ methods, but there's no exact correspondence with a limited set of papers. \end{tabular} \\ \vspace{1.5mm}
        2 & \begin{tabular}[c]{@{}l@{}}\textbf{Minor Similarity:} The LLM proposal bears very slight resemblance  with some existing \\ papers. Mostly novel. \end{tabular} \\ \vspace{1.5mm}
        1 & \begin{tabular}[c]{@{}l@{}}\textbf{Original:} The LLM proposal is completely novel. \end{tabular} \\
        \bottomrule
    \end{tabular}
    \caption{Scoring rubric shared with experts to evaluate similarity of LLM-generated research proposals with prior work.}
    \label{tab:similarity-scores}
\end{table*}

Our analysis reveal a concerning pattern
wherein a significant portion of LLM-generated research ideas
appear novel on the surface
but are actually skillfully plagiarized
in ways that make their originality difficult to verify.
The case study presented in \S\ref{subsec:case-study} supports this thesis, examining a research proposal (an exemplar in \citep{si2024can}) that appears to be skillfully plagiarized from an existing paper.
Our analysis in \S\ref{sec:discussion} also 
finds that LLM-generated research content 
is less diverse 
and 
follows more predictable patterns. 
Our preliminary investigation suggests these patterns might be detectable through basic classification methods, potentially helpful in flagging content for additional review, though more research is needed to develop robust detection approaches.

While we do not recommend wholesale dismissal of LLM-generated research,
our findings suggest that they  
may not be as novel as 
previously thought, and additional scrutiny is warranted.
The sophisticated nature of the plagiarism
we uncover suggests 
that widespread adoption of these tools 
could significantly impact the peer review process, 
requiring (already overwhelmed) reviewers 
to spend additional time searching 
for potential content misappropriation.

\section{Related Work}
\label{sec:related}

\paragraph{Novelty of LLM Generated Research.} 
Evaluating 
novelty of 
generated proposals 
typically follow two paths: automated evaluation using LLMs themselves \citep{lu2024ai} or human expert review \citep{li2024chain, baek2024researchagent, li2024mlr, wang2023scimon, yang2023large, li2024learning, weng2024cycleresearcher}, often 
conducted 
through 
a small group of known experts. The scope and detail of generated outputs varies across studies, with some work focusing on concise proposals \citep{wang2023scimon,yang2023large}, while other approaches generate more detailed proposals or complete research papers \citep{si2024can,lu2024ai}. 
A recent large-scale study improves how we evaluate LLMs' ability to generate research proposals, with experts from multiple institutions reviewing $81$ LLM-generated proposals \citep{si2024can}. 
This study reveals several key findings. First, LLMs demonstrate limited ability in evaluating research ideas. Second, through a carefully controlled methodology that generates and ranks multiple candidates using significant computational resources, the study finds a surprising result: human participants judge LLM-generated proposals as more novel than human-written ones.

\paragraph{Automated Plagiarism Detection Tools.} \label{para:rel-automated-plag-tools}
Several studies integrate LLMs with academic search engines like Semantic Scholar to detect and filter out potential plagiarism in LLM-generated research ideas \citep{si2024can, lu2024ai, li2024chain}. The typical methodology  involves extracting keywords from titles and abstracts using LLMs, querying these through Semantic Scholar's API, and performing one-to-one comparisons between retrieved papers and the generated ideas. While not specifically designed for plagiarism detection, OpenScholar \citep{asai2024openscholar} is a specialized retrieval-augmented LM that leverages a database of $45$ million open-access papers with $237$ million passage embeddings. Its retrieval mechanism combines nearest-neighbor search over passage embeddings, keyword-based search through Semantic Scholar API, and academic web search results. Given this sophisticated multi-source retrieval system and its vast database of passage embeddings, OpenScholar could potentially serve as a powerful tool for embedding-based plagiarism detection. Traditional text-matching tools like Turnitin are also commonly used.

\paragraph{LLMs in Research Tasks.} Recent research has explored LLMs' capabilities in predicting experimental outcomes \citep{luo2024large}, conducting research experiments \citep{huang2023benchmarking, tian2024scicode}, paper reviewing \citep{weng2024cycleresearcher, zeng2023scientific}, and related work generation \citep{hu2024hireview}. We refer the reader to \citet{luo2025llm4sr} for a survey on this topic.

\paragraph{Creativity in AI.} Our work connects to studies on AI creativity \citep{ismayilzada2024creativity}. 
For instance, studies on LLM poetry generation reveal that while humans struggle to distinguish between AI-generated and human-written poems \citep{porter2024ai},
the outputs contain substantial verbatim matches with web text \citep{djsearch}, suggesting limited original creativity.

\section{Background: Generating Proposals}
\label{sec:background}

Here, we elaborate on the methodology
used to generate research proposals \citep{si2024can},
with particular attention to the plagiarism detection module, 
as understanding it is central to our discussion.

The research proposal generation process consists of six sequential steps,
with Claude $3.5$ Sonnet \citep{claude35sonnet} being the backbone LLM. 
For a given topic, the first 
step uses a retrieval-augmented generation (RAG) system 
to retrieve and rank 
relevant papers 
using the Semantic Scholar API. %
The second step generates initial seed ideas 
using these retrieved papers, 
while the third step 
involves 
deduplication using text embeddings, retaining about $5\%$ of the original ideas. 
The fourth step expands these seed ideas into detailed project proposals containing a title, problem statement, motivation, proposed method, experiment plan, test cases, and a fall back plan.
The fifth step implements a Swiss tournament system 
where proposals compete in pairwise comparisons over five rounds 
to identify the strongest candidates. 
We refer readers to \cite{si2024can} for additional details on these steps.

The final step, 
most relevant to our work, 
attempts 
to detect potential plagiarism through Semantic Scholar Augmented Generation (SSAG). 
First, an LLM iteratively 
generates queries 
for the Semantic Scholar API 
to find papers similar to the proposal's content,
using results from previous iterations as context to inform each new query.
This search process continues 
either until they collect $100$ papers 
or reach $10$ iterations, 
whichever comes first. 
The retrieved papers are then scored by an LLM 
based on their relevance, 
narrowing down to $10$ most similar papers. 
Further, Claude $3.5$ Sonnet 
performs pairwise comparisons 
between the proposal and each of these top $10$ papers. 
The LLM 
determines if the proposal and the retrieved paper are substantially similar, 
discarding the proposal if they are. 
This process removes \emph{only} about $1\%$ of the generated proposals \citep{si2024can}.

Similar 
detection approaches
have been implemented in other research agents \citep{li2024chain,lu2024ai}, 
with minor variations in prompting strategies 
and parameters---\citet{li2024chain} and \citet{lu2024ai} query Semantic Scholar $3$ and $10$ times to search for similar papers, respectively. 

For our study's implementation, \citet{si2024can}'s method is slightly modified to optimize computational costs while maintaining effectiveness.  Instead of generating $4000$ ideas per research topic,  we generate $500$ ideas per topic.  This reduces our initial pool from around $200$ unique ideas per topic (as in the original paper \citep{si2024can})  to $138$,  resulting in significant computational cost savings  while only reducing unique ideas by $31\%$.
This non-linear relationship between generated and unique ideas can be explained by the fact that 
as research agents generate more ideas, 
the percentage of unique ideas decreases and eventually plateaus, 
offering diminishing returns \citep{si2024can}.
All other components of the pipeline,  including the plagiarism detection and ranking,  are exactly the same as \citet{si2024can}'s original implementation.

The choice to use this methodology \citep{si2024can}
for generating research proposals has several motivations. 
First, it is representative of prompt engineering techniques 
used across various research idea generation systems \citep{li2024chain, lu2024ai, baek2024researchagent, li2024mlr, wang2023scimon, yang2023large, weng2024cycleresearcher}. 
Second, as discussed in \S\ref{sec:related}, 
this study conducts the most comprehensive evaluation to date, 
and finds that human participants judge LLM-generated research 
more 
novel than those 
by humans. 
Third, this method uses minimal prompt engineering. Other approaches like \citet{li2024chain} use sophisticated prompt engineering that function as ``natural language programs,'' potentially incorporating human creativity. By keeping prompts simple, we can better assess LLMs' raw capabilities.

\section{Experimental Design}
\label{sec:experiments}

We present our expert evaluation design for detecting plagiarism in LLM-generated research documents and our evaluation methodology for automated plagiarism detection tools.

\subsection{Expert Evaluation Design}
\label{subsec:expert-evaluation-design}

Our experimental setup involves generating 
five research proposals for each of the twelve topics listed in Table~\ref{tab:research-topics}. 
These NLP topics are determined 
by asking participants 
to describe their areas of expertise, 
ensuring evaluators have  
in-depth familiarity with 
the current literature for detecting potential plagiarism. 
Twelve participants each evaluate $3$ out of $5$ proposals that align with their domain knowledge, resulting in $3\times12=36$ total proposals.

In addition to these $36$ proposals, 
we include fourteen previously-showcased exemplars---four research proposals from \citet{si2024can} 
and ten research papers from \citet{lu2024ai}, 
bringing the total to $50$ research documents. 
Our study includes $13$ experts in total (including the first author of this work): $12$ experts evaluate the 36 proposals, and $2$ experts evaluate the $14$ exemplars, 
with one expert being common among both evaluation tasks.
We use convenience sampling 
to recruit participants who are actively conducting research in NLP 
through our professional networks. 
Our participant pool comprises experts from $5$ universities
and $2$ industrial labs,
with $69\%$ being Ph.D. students or recent graduates
and $31\%$ being associate researchers in industrial or academic labs.\footnote{Our experts are affiliated with University of Washington, Carnegie Mellon University, Harvard University, University of Southern California, Indian Institute of Science, Allen Institute for Artificial Intelligence (Ai2)
and Adobe, Inc.}
Unlike prior studies 
where participants assess LLM-generated research documents 
without suspecting plagiarism, 
our participants are instructed to actively search 
for potential overlaps 
with prior work.

Participants are asked to only 
consider papers available until April $2024$, 
corresponding to Claude $3.5$ Sonnet's training cutoff date.
Using a consistent evaluation rubric (see Table~\ref{tab:similarity-scores}), participants assign similarity scores from $1$-$5$ to each proposal. We focus on documents scoring $4$ or $5$, as these represent clear cases of content misappropriation---score $5$ indicates direct copying with one-to-one mapping to existing methods, while score $4$ indicates significant borrowing from prior work without attribution.
These scores reflect methodological overlap considered plagiarism in academic publishing, 
as they represent either wholesale adoption of existing methods (score $5$) 
or substantial uncredited incorporation of others' technical contributions (score $4$). 
Lower scores represent more ambiguous cases of potential similarity. 

When evaluating plagiarism by humans, intent is considered important---whether the copying was deliberate deception or unintentional failure to attribute sources. However, assessing LLM intent is not feasible. The source papers we identify were likely present in the model's training data (as we ask participants to only report papers before Claude $3.5$ Sonnet's cutoff date), but LLMs cannot reliably report which sources influenced their outputs or how they synthesized information during generation. 
Therefore, we evaluate based on observable methodological similarities rather than attempting to determine intent. For all documents with scores $4$ and $5$, we email source paper authors for verification and adjust scores based on their feedback. Since some authors were unreachable, we report both verified claims and total claims separately.

 Unlike conventional human studies 
 that aim for objectivity 
 through unbiased instructions 
 and experimental design, 
 our expert evaluation design 
 derives objectivity primarily 
 from author verifications and 
 the inherently verifiable nature 
 of plagiarism claims---readers 
 can independently examine both the source 
 and generated works through our open-sourced results. 
 The instructions shared with participants are shown in Table~\ref{tab:expert-instructions}. We discuss some limitations of our expert evaluation design in \S\ref{para:lim-expert-evaluation}.

\subsection{Evaluating Plagiarism Detectors}
\label{subsec:automated-plag-detection-exp}

To systematically evaluate automated plagiarism detection tools, 
we require a dataset of plagiarized research documents paired with their original source papers. 
While our expert evaluation study uncovers instances of plagiarism, 
the sample size is too small to comprehensively test automated tools. 
We therefore create a synthetic test set by 
generating plagiarized research proposals from papers 
retrieved during the literature review step of the baseline method \citep{si2024can}. 

For the twelve research topics listed in Table~\ref{tab:research-topics}, 
we select $40$ papers per topic, 
creating a test set of $480$ papers.
We then use GPT-4o~\citep{openai2024gpt4ocard} 
to generate plagiarized versions of these papers 
by skillfully paraphrasing the paper's details
to avoid detection.
We choose GPT-4o for this task because Claude $3.5$ Sonnet abstains from plagiarism requests. 
The specific prompt used for this process is detailed in Table~\ref{tab:generation-prompt}.

This synthetic testing 
approach poses a significantly easier challenge 
than detecting plagiarism in proposals 
generated by sophisticated research agents. 
Since we explicitly instruct GPT-4o 
to plagiarize from a single paper,
novel elements in these deliberately plagiarized proposals 
are likely more limited 
than those produced by systems designed to generate novel research. 
Therefore, the performance of automated detection systems on our test set 
likely overestimates their ability to detect more subtle forms of plagiarism 
in LLM systems designed to generate novel research content.

Successful plagiarism detection 
requires two steps: 
first retrieving potentially plagiarized source papers, 
and then determining whether the retrieved papers 
are substantially similar to the proposal in question. 
We design experiments to evaluate these 
components both separately and together across three approaches. 
The first approach uses two LLMs (GPT-4o and Claude 3.5 Sonnet) 
in three distinct scenarios: 
(a) oracle access, wherein we provide an LLM with 
both the proposal and its source paper
and evaluate its ability to detect plagiarism,
(b) parametric knowledge testing, 
which examines LLMs' ability to both retrieve and determine 
similarity using only their training data without external tools, 
and (c) Semantic Scholar Augmented Generation (SSAG), described in Section~\ref{sec:background}, 
which explicitly 
separates these steps by first retrieving papers through Semantic Scholar 
and then determining similarity.\footnote{For SSAG, we use the following hyperparameters: maximum $50$ papers and $5$ iterations.}

The other two approaches, described in \S\ref{para:rel-automated-plag-tools}, are OpenScholar \citep{asai2024openscholar} (prompt in Table~\ref{tab:openscholar-prompt}), an academic search system with a database of $45$ million papers and sophisticated retrieval mechanisms, and Turnitin \cite{turnitin}, a widely-used commercial plagiarism detection service. Due to the manual effort required for submission and analysis, we limit our testing with these tools to $3$ papers per research topic, totalling $36$.

\section{Results}
\label{sec:results}

Our code and detailed findings, including all evaluated proposals and author verifications, are available at \url{https://github.com/tarun360/AI-Papers-Plagiarism/}.

\subsection{Expert Evaluation Results}
\label{subsec:expert-evaluation-results}

\begin{table}[t]
    \centering
    \begin{tabular*}{\columnwidth}{@{\extracolsep{\fill}} l cc}
        \toprule
        \textbf{Score} & \textbf{Total Claims} & \textbf{Verified} \\
        & \textbf{(\%)} & \textbf{(\%)} \\
        \midrule
        5 & 18.0\% (9/50) & \textbf{14.0\% (7/50)} \\
        4 & 18.0\% (9/50) & \textbf{10.0\% (5/50)} \\
        3 & 32.0\% (16/50) & 8.0\% (4/50) \\
        2 & 28.0\% (14/50) & 4.0\% (2/50) \\
        1 & 4.0\% (2/50) & 0.0\% (0/50) \\
        \bottomrule
    \end{tabular*}
    \caption{Distribution of similarity scores for LLM generated proposals. Considering scores $4$ and $5$ as instances of plagiarism, $24.0\%$ of examined proposals 
    ($36.0\%$ if including unverified claims) 
    are plagiarized.
    We only verify claims for proposals with initial scores of $4$ and above, 
    therefore the total number of verified proposals is less than 50.
    }
    \label{tab:plagiarism-distribution}
\end{table}

Our expert evaluation reveals 
plagiarism in LLM-generated research documents, 
as shown in Table~\ref{tab:plagiarism-distribution}. 
Similarity scores 
for verified claims highlight substantial content misappropriation at various levels of severity. 
Since our evaluation is limited by our participants' time constraints 
and the laborious manual effort required for thorough plagiarism checks, 
our results likely represent a lower bound on actual plagiarism rates. 
Interestingly, 
of the four exemplars presented in \citet{si2024can}, 
one received a similarity score of $5$ 
and another received a score of $4$, 
while among the ten exemplars in \citet{lu2024ai}, 
two received scores of $5$ and one received score $4$---all
of these ratings are cross-verified 
by the source papers' authors.
It is crucial 
to note 
the documents 
that receive similarity
scores of $5$ in our evaluation
passed through SSAG plagiarism detection checks,
indicating serious limitations of such systems (elaborated in §\ref{subsec:automated-plag-exp-results}).

We also evaluate OpenScholar and Turnitin
on research documents with verified scores of $4$ or above. 
We provided OpenScholar with each document's title, problem statement, motivation section, and methodology section using the prompt shown in Table \ref{tab:openscholar-prompt}. Among the $4$-$5$ papers that OpenScholar suggested as related works for each document, the actual source paper appeared only in one case.
Turnitin did not identify the original paper in any case.
We further analyze these findings through a detailed case study in \S\ref{subsec:case-study}.

\subsection{A Case Study}
\label{subsec:case-study}

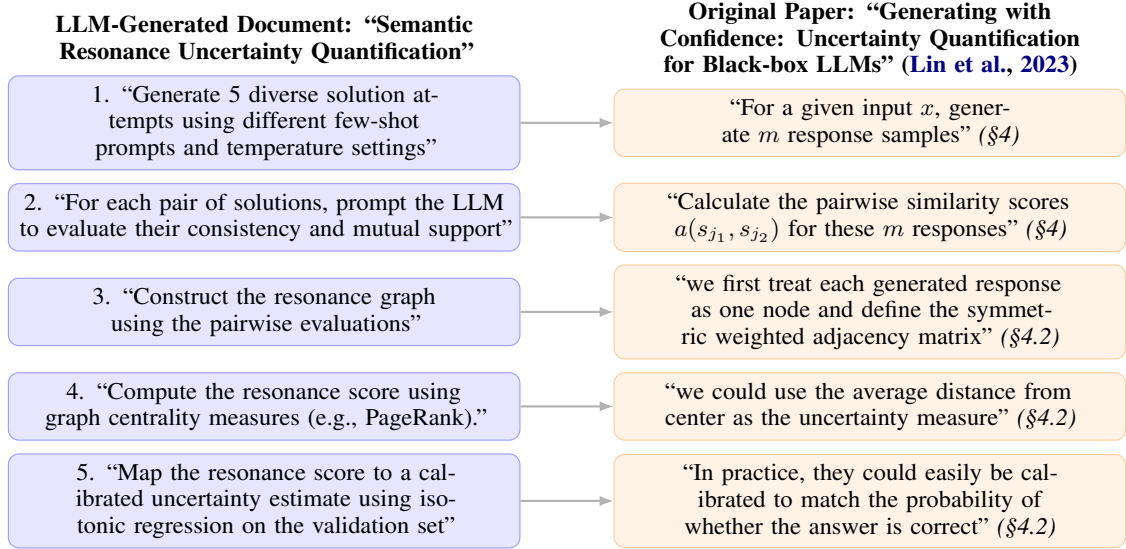
\begin{figure*}[t]
    \centering
    \begin{tikzpicture}[
        box/.style={draw, rounded corners, text width=6.5cm, minimum height=2em, align=center, font=\small},
        leftbox/.style={box, fill=blue!10, draw=blue!40},
        rightbox/.style={box, fill=orange!10, draw=orange!40},
        arrow/.style={->, >=latex, thick, draw=gray!60},
        title/.style={font=\small\bfseries, text width=6.5cm, align=center}
    ]
        \node[title] (t1) at (-4, 3.35) {LLM-Generated Document: ``Semantic \\Resonance Uncertainty Quantification''};
        \node[title] (t2) at (4, 3.35) {Original Paper: ``Generating with \\Confidence: Uncertainty Quantification for Black-box LLMs'' \citep{lin2023generating}};
        
        \node[leftbox] (l1) at (-4, 2.25) {1. ``Generate 5 diverse solution attempts using different few-shot prompts and temperature settings''};
        \node[leftbox] (l2) at (-4, 1) {2. ``For each pair of solutions, prompt the LLM to evaluate their consistency and mutual support''};
        \node[leftbox] (l3) at (-4, -0.25) {3. ``Construct the resonance graph using the pairwise evaluations''};
        \node[leftbox] (l4) at (-4, -1.5) {4. ``Compute the resonance score using graph centrality measures (e.g., PageRank).''};
        \node[leftbox] (l5) at (-4, -2.75) {5. ``Map the resonance score to a calibrated uncertainty estimate using isotonic regression on the validation set''};
        
        \node[rightbox] (r1) at (4, 2.25) {``For a given input $x$, generate $m$ response samples'' \textit{(\S4)}};
        \node[rightbox] (r2) at (4, 1) {``Calculate the pairwise similarity scores $a(s_{j_1}, s_{j_2})$ for these $m$ responses'' \textit{(\S4)}};
        \node[rightbox] (r3) at (4, -0.25) {``we first treat each generated response as one node and define the symmetric weighted adjacency matrix'' \textit{(\S4.2)}};
        \node[rightbox] (r4) at (4, -1.5) {``we could use the average distance from center as the uncertainty measure'' \textit{(\S4.2)}};
        \node[rightbox] (r5) at (4, -2.75) {``In practice, they could easily be calibrated to match the probability of whether the answer is correct''  \textit{(\S4.2)}};
        
        \foreach \i in {1,...,5} {
            \draw[arrow] (l\i) -- (r\i);
        }
    \end{tikzpicture}
    \caption{Visual mapping between an LLM-generated research document (an exemplar in \citep{si2024can}) and a published paper \citep{lin2023generating}, showing a direct correspondence in their proposed methodologies. Each element of the proposed method has a corresponding match in the source paper, suggesting sophisticated rewording rather than novel contribution. This pair receives a similarity score of $5$ in our expert evaluation, which is verified by the authors of the source paper.}
    \label{fig:plagiarism-example}
\end{figure*}

To illustrate the nature of  
plagiarism in LLM-generated research documents, 
we examine a proposal (an exemplar in \citep{si2024can}) 
titled ``Semantic Resonance Uncertainty Quantification'' 
(readers can find this document in their paper). 
This document receives a similarity 
score of $5$ in our evaluation, 
with direct correspondence to an existing published paper, 
``Generating with Confidence: Uncertainty Quantification for Black-box Large Language Models'' \citep{lin2023generating}. 
The authors of the original paper 
confirm our plagiarism assessment 
after reviewing the proposal in question.

As shown in Figure~\ref{fig:plagiarism-example}, 
the proposed methodology exhibits 
a clear one-to-one mapping with the original paper. 
Each component of the LLM-generated 
proposal corresponds to specific sections in the source paper,
albeit with skillfully reworded descriptions. 
The proposal 
proposes the same technical approach 
while using different terminology 
(e.g., ``resonance graph'' instead of ``weighted adjacency matrix'') 
and restructuring the presentation. 
This may be interpreted as adversarial behavior, 
where the LLM has 
learned to disguise existing work as novel research 
through careful rewording. 
Notably, expert reviewers
in \citet{si2024can}'s study 
do not identify this plagiarism, 
likely because 
their evaluation focuses on assessing novelty and feasibility
rather than actively searching for potential sources of plagiarism. 
In contrast, our study's participants, 
operating under different situational logic \citep{popper2013poverty,hoover2016situational} 
that presumes potential plagiarism, 
are able to identify the original paper. 
This case-study, 
along with results 
of expert evaluation presented in \S\ref{subsec:expert-evaluation-results}, 
showcases the limitations of previous human evaluations of 
LLM generated research ideas---without a skeptical eye for plagiarism, 
experts may be fooled into considering LLM generated research ideas as novel.
Additional case studies examining 
plagiarism and similarity in AI-generated research 
from \citep{lu2024ai} and \citep{yamada2025ai} 
are available in \S\ref{appendix:additional-case} and \S\ref{appendix:workshop-paper-case} respectively.

\subsection{Performance of Plagiarism Detectors}
\label{subsec:automated-plag-exp-results}

\begin{table}[ht]
    \centering
    \small
    \renewcommand{\arraystretch}{1.3}
    \begin{tabular*}{\columnwidth}{@{\extracolsep{\fill}} llr}
        \toprule
        \multicolumn{2}{c}{\textbf{Method}} & \multicolumn{1}{l}{\textbf{Accuracy}} \\
        \midrule
        \multirow{3}{*}{Claude 3.5 Sonnet} & Oracle access & 88.8\% \\
         & Parameteric Knowledge & 1.3\% \\
         & SSAG & 51.3\% \\
        \midrule
        \multirow{3}{*}{GPT-4o} & Oracle access & 89.0\% \\
         & Parameteric Knowledge & 32.7\% \\
         & SSAG & 68.5\% \\
        \midrule
        \multicolumn{2}{l}{OpenScholar} & 0\% \\
        \midrule
        \multicolumn{2}{l}{Turnitin} & 0\% \\
        \bottomrule
    \end{tabular*}
    \caption{Comparing performance of automated plagiarism detectors, including $2$ LLMs in three scenarios (oracle access, parametric knowledge, SSAG), OpenScholar, and Turnitin.}
    \label{tab:model-comparison}
\end{table}

To evaluate automated plagiarism detection methods, we test LLMs in three scenarios (oracle access, parametric knowledge, and SSAG), OpenScholar, and Turnitin. Table~\ref{tab:model-comparison} presents their performance metrics across these settings. Even in our simplified test scenario with deliberately plagiarized proposals, detection accuracy remains remarkably low across all methods.

The oracle access setting, 
where models are given both 
the ground truth paper and plagiarized proposal, 
yields the highest accuracy ($88.8\%$ for Claude 3.5 Sonnet and $89.0\%$ for GPT-4o). 
However, this represents an idealized scenario rarely possible in practice, 
where the source of potential plagiarism is unknown. 
When testing models' parametric knowledge 
without access to external tools, 
GPT-4o achieves notably higher accuracy ($32.7\%$) 
compared to Claude 3.5 Sonnet ($1.3\%$), 
likely because GPT-4o is used to generate these plagiarized proposals in the first place.

The SSAG approach,
a common plagiarism detection method 
in research document generation systems \citep{si2024can,lu2024ai,li2024chain}, 
attains moderate performance ($51.3\%$ for Claude, $68.5\%$ for GPT-4o).
The considerable gap between 
SSAG and oracle access performance  
indicates that retrieving relevant papers, not determining similarity, 
is the bottleneck.

SSAG's inadequate performance is particularly concerning for research agent systems relying on it \citep{si2024can,lu2024ai,li2024chain}, especially given our test scenario represents a much easier challenge---our proposals are deliberately plagiarized from single papers, while research agent systems are designed to produce allegedly novel content. (An illustrative example of 
successful plagiarism detection 
using the SSAG
from our synthetic dataset is provided in Appendix~\ref{appendix:successful-detection}.)

Both OpenScholar and Turnitin, 
tested on a smaller subset of proposals ($3$ per topic), 
fail to detect any instances of plagiarism. 
These results, 
combined with significant plagiarism found in our expert evaluation (\S\ref{subsec:expert-evaluation-results}), 
underscore that current automated plagiarism detection methods are inadequate 
for identifying content misappropriation in LLM-generated research documents.

\section{Discussion}
\label{sec:discussion}

\paragraph{Comparison with Human-Written Papers.}
In principle, 
it would be useful
to conduct a similar analysis of human-written papers to estimate rates of plagiarism in them.
However, given the labor-intensive nature of 
finding plagiarism in LLM-generated papers, 
our main study focuses on analyzing LLM-generated content alone. 
As a proxy to the expert-led study on human-written papers, %
we conduct an automated analysis 
to extract signs of plagiarism from peer reviews of human-written papers.
We use existing peer reviews from the PeerRead dataset \citep{kang-etal-2018-dataset} 
for human-written papers from NeurIPS 2017 ($499$ papers), ACL 2017 ($123$ papers), ICLR 2017 ($349$ papers), and CoNLL 2016 ($19$ papers). 
For NeurIPS 2017,
we sampled $499$ papers from the larger set due to computational constraints.
We use Claude 3.7 Sonnet to extract 
potential concerns of plagiarism mentioned in peer reviews, 
using the prompt shown in Table~\ref{tab:peer-review-prompt}.
We emphasize that the LLM is used only to extract information about reviewer claims, 
not to make plagiarism judgments. 
As shown in Table~\ref{tab:human-plagiarism}, 
the estimated plagiarism rate in human-authored papers is significantly lower than the $24\%$ we found in LLM-generated papers. 
This stark difference reinforces our finding that LLM-generated research exhibits 
substantially higher rates of content misappropriation than traditional human-authored work.

\begin{table}[t]
    \centering
    \small
    \begin{tabular*}{\columnwidth}{@{\extracolsep{\fill}} l ccc}
        \toprule
        \textbf{Conference} & \textbf{Score $4$} & \textbf{Score $5$} & \textbf{Plagiarism} \\
        & & & \textbf{rate (scores $4$+)} \\
        & \textbf{(\%)} & \textbf{(\%)} & \textbf{(\%)} \\
        \midrule
        ACL 2017 & $0.8\%$ & $0\%$ & $0.8\%$ \\
        ICLR 2017 & $4.0\%$ & $2.3\%$ & $6.3\%$ \\
        CoNLL 2016 & $5.3\%$ & $0\%$ & $5.3\%$ \\
        NeurIPS 2017 & $1.8\%$ & $0\%$ & $1.8\%$ \\
        \bottomrule
    \end{tabular*}
    \caption{We estimate plagiarism rates in human-written papers by automatically extracting signs of plagiarism from peer reviews. 
    We find that the overall plagiarism rate is significantly lower than the $24\%$ found in LLM-generated proposals.}
    \label{tab:human-plagiarism}
\end{table}

\begin{figure}
    \centering
    \includegraphics[width=0.475\textwidth]{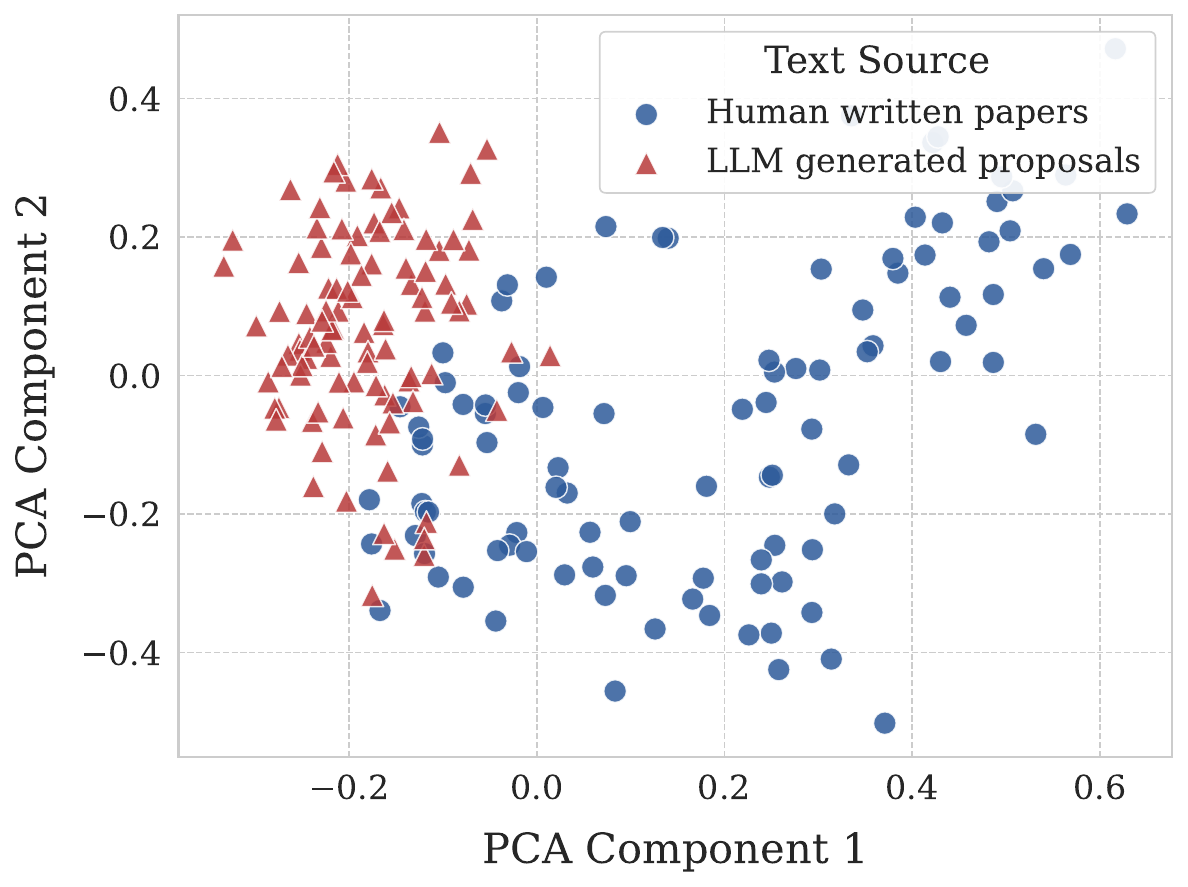}
    \caption{PCA projection of concatenated title and abstract embeddings for human-written papers and LLM-generated proposals on the topic 
    ``Novel AI-assisted formal proof generation methods''. LLM-generated proposals occupy a more confined region, indicating less diversity in outputs.}
    \label{fig:scatter}
\end{figure}

\paragraph{Limited Diversity in Proposals.}
\label{para:titles-clustering} 
Prior research notes limited diversity in LLM-generated research proposals \citep{si2024can}, 
which our analysis confirms. 
For each research topic in Table~\ref{tab:research-topics}, 
we analyze concatenated titles and abstracts of $100$ human-written papers and $100$ LLM-generated proposals.\footnote{For these experiments, we modify \citet{si2024can}'s proposal generation to output title and abstract format (instead of the original structure described in \S\ref{sec:background}) to enable direct comparison with human-written papers.\label{footnote:modified-proposal-generation}}\ 
Using the all-MiniLM-L6-v2 sentence transformer \citep{reimers2020making}, 
we generate embeddings for these concatenated titles and abstracts.
Calculating cluster spread via mean squared distance from centroids, 
we find that despite aggressive deduplication strategies \citep{si2024can}, 
proposals generated by both LLMs are more tightly clustered---the ratio of LLM to human cluster 
spreads is $0.81$ for proposals generated by Claude 3.5 Sonnet and $0.72$ for proposals generated by GPT-4o across topics.\footnote{Due to deduplication and filtering in \citep{si2024can}, some topics have $< 100$ proposals, resulting in $2370$ samples for Claude 3.5 Sonnet and $2019$ samples for GPT-4o rather than the maximum of $2400$ ($200$ samples $\times$ $12$ topics).\label{footnote:less-samples}}\
Figure~\ref{fig:scatter} illustrates this pattern, 
showing that LLM proposals occupy a more confined region in the embedding space. 
While this analysis is limited since titles and abstracts only reflect broad research directions rather than specific ideas,
the clear clustering pattern suggests current LLM systems explore a narrower range of research directions compared to humans, 
undermining their utility for scientific innovation.

\paragraph{Proposals Are Easily Distinguishable.}
\label{para:titles-classification} 
Beyond being tightly clustered, 
LLM-generated proposals also appear to follow distinctive patterns in their research directions. 
In a preliminary analysis using $100$ human-written papers
and $100$ LLM-generated proposals per topic from Table~\ref{tab:research-topics}, 
we train separate binary classifiers to distinguish between human-written papers and LLM-generated 
proposals from Claude 3.5 Sonnet, 
and between human-written papers and proposals from GPT-4o.\footref{footnote:modified-proposal-generation}\ We use a simple logistic regression model
with all-MiniLM-L6-v2 sentence transformer \citep{reimers2020making} features on a $60$:$40$ train-test split.
The classifiers achieve $93.9\%$ accuracy (Claude 3.5 Sonnet vs. human) and $92.6\%$ accuracy (GPT-4o vs. human).\footref{footnote:less-samples}\
While this basic classification experiment has limitations as titles and abstracts only capture broad research directions, 
it provides initial evidence that LLM-generated research follows predictable patterns. 
However, developing reliable detection methods will require significant further research 
and is beyond the scope of this work.

\paragraph{Citing Relevant Work Is Challenging.} 
One might argue that 
requesting LLMs 
to provide citations while generating research documents could mitigate plagiarism concerns. 
In our preliminary experiments, 
we manually examine several cases 
where we ask LLMs to generate proposals with citations. 
However, 
the LLMs typically reference a few well-known papers, 
raising concerns about citation accuracy. 
A comprehensive evaluation of 
citation quality 
would require expert verification of each citation and, 
more importantly, 
identification of relevant citations that were omitted, 
making large-scale testing expensive. 
This limitation aligns with broader findings in the field---\citet{asai2024openscholar} 
find that $78$-$90\%$ of papers cited by non-retrieval augmented LLMs are hallucinated. 
While recent works \citep{asai2024openscholar, Gao2023EnablingLL, Qian2024OnTC} attempt to improve citation generation, 
their findings suggest that LLMs struggle with accurate citations, 
leading researchers to rely on RAG 
and other external tools for citation support.

\paragraph{Implications for Academic Publishing.} 
Presence of 
considerable plagiarism 
in LLM-generated 
research documents suggests 
that widespread adoption of these 
tools could lead to an increase in 
publications 
with improper citations or inadvertent plagiarism. 
Furthermore, 
if researchers execute LLM-generated proposals 
and submit their work to conferences and journals, 
the sophisticated nature of this plagiarism 
would require conference and journal reviewers 
to spend considerably more time searching for potential content misappropriation, 
adding to an already heavy reviewer workload.

\paragraph{Future Research Directions.} 
Developing ways 
to identify 
candidate source papers 
is one of the most pressing future research direction, 
as current automated detection methods are inadequate 
and manual evaluation by domain experts is both time-consuming and laborious.
Future work could also explore post-training strategies that could help reduce plagiarism in LLM-generated research content.
Lastly, 
future studies could examine 
whether LLM-generated 
content 
directly copy, or significantly borrow,
content from copyrighted materials, possibly through the lens of fair-use doctrine \citep{patterson1992understanding, balaji2024fair}.

\section{Conclusion}

We conducted the first systematic study of plagiarism in LLM-generated research documents. Through an expert-led evaluation, we discovered significant levels of plagiarism in these documents. We demonstrated the inadequacy of current automated plagiarism detection methods, 
while our case studies revealed sophisticated forms of content misappropriation that passed through multiple layers of filtering and expert review. 
Our findings raise important concerns about potential wide-spread use of LLMs for research ideation and highlights the need for better plagiarism detection methods.

\section{Limitations} 
\label{sec:limitations}

\paragraph{Automation Challenges.}
A key limitation identified in our study stems 
from the challenge of automating the 
detection of original papers that may have been plagiarized. 
Currently, no reliable automated systems exist for this task, 
necessitating reliance on human expertise. 
This manual process is time-intensive, making it a critical bottleneck.

\paragraph{Constraints Related to Expert Evaluation.} \label{para:lim-expert-evaluation}
Our expert evaluation design has some limitations. 
First, we ask our human participants to presume adversarial plagiarism and actively search for it. 
This may introduce confirmation bias, 
leading them to give higher similarity scores in order to complete their task. 
We indeed find $4$ instances 
where source-papers authors over-turned the score downwards 
with differences of $2$ or more. 
However, our (author-verified) scores already account for these adjustments, 
and our findings remain independently verifiable 
through our open-sourced results. 
Second, we provide our human-participants with $5$ research proposals and ask them to choose any $3$. 
While this gives flexibility to our participants, 
since even in a single topic there can be sub-topics that participants may not be familiar with, 
it may introduce sampling bias.

\paragraph{Computational Parameter Reductions.} 
We reduce the quantity of certain hyperparameters to optimize computational costs. 
First, as detailed and motivated in \S\ref{sec:background}, 
we decrease the number of candidate proposals generated per topic, 
resulting in approximately $31\%$ fewer unique ideas compared to the baseline method \citep{si2024can}. 
Although this reduction might marginally affect the quality of research proposals relative to \citet{si2024can}, 
the validity of our findings remains, 
particularly given that we identify plagiarism even in the LLM-generated proposals showcased in their original work. 
Second, in our automated plagiarism detection experiments (\S\ref{subsec:automated-plag-detection-exp}), 
we limit Semantic Scholar queries to a maximum of $5$ iterations, 
fewer than the $10$ iterations employed in \citet{si2024can,lu2024ai}, but more than the $3$ iterations in \citet{li2024chain}. 
However, as elaborated in \S\ref{subsec:automated-plag-detection-exp}, 
our synthetic dataset presents a simpler challenge than detecting plagiarism in actual research proposal generation pipelines
(note that for generating proposals in our expert evaluation study, we maintain the same number of Semantic Scholar queries as \citet{si2024can} in the plagiarism filtering step).

Despite these limitations, 
we believe that our study highlights a critical concern, 
and adds nuance to the 
ongoing discourse about the role of LLMs in scientific research.

\section*{Acknowledgements}

We are grateful to 
Chenglei Si for publicly releasing the code to generate research proposals
and kindly answering our questions.
We sincerely thank all participants in our evaluation study for their valuable time and insights, including Anirudh Ajith, Siddhant Arora, Rachit Bansal,  Miroojin Bakshi, Kirti Bhagat, Brihi Joshi, Navreet Kaur, Rishubh Parihar, Ananya Sai, Prarabdh Shukla, Apoorv Umang and Kinshuk Vasisht.  We are grateful to Saksham Rastogi and Kinshuk Vasisth for their feedback on our work. DP is grateful to Adobe Inc., Schmidt Sciences, National Payments Corporation of India (NPCI), Kotak AI Center, and Pratiksha Trust for sponsoring his group's research. The search icon in Figure \ref{fig:expert-evaluation-setup} is from \citet{rsetiawan2025search}.

\bibliography{acl25}

\appendix
\newpage 

\begin{table*}[t]
    \centering
    \small
    \renewcommand{\arraystretch}{1.4}  %
    \begin{tabular}{>{\raggedright\arraybackslash}p{0.95\textwidth}}
        \toprule
        \textbf{Research Topics} \\
        \midrule
        Improving long context capabilities of large language models. \\[1ex]
        Evaluating abstention capabilities and techniques for language models. \\[1ex]
        Evaluating geographical and cultural bias in large language models. \\[1ex]
        Novel methods to improve trustworthiness and reduce hallucinations of large language models. \\[1ex]
        Novel methods for mechanistic interpretability of large language models. \\[1ex]
        Novel methods to add speech and audio processing capabilities into large language models. \\[1ex]
        Novel AI-assisted formal proof generation methods. \\[1ex]
        Human-centric evaluation of large language models, and development of language technologies from a social perspective. \\[1ex]
        Novel techniques and metrics for evaluating machine translation systems. \\[1ex]
        Novel methods to understand neural scaling laws for large language model training. \\[1ex]
        Novel methods to improve inference performance of large language models. \\[1ex]
        Novel methods for developing and evaluating large language model based personas. \\
        \bottomrule
    \end{tabular}
    \caption{Research topics in natural language processing used for generating LLM proposals and matching with expert evaluators' domain expertise.}
    \label{tab:research-topics}
\end{table*}

\section{Case Study 2}\label{appendix:additional-case}

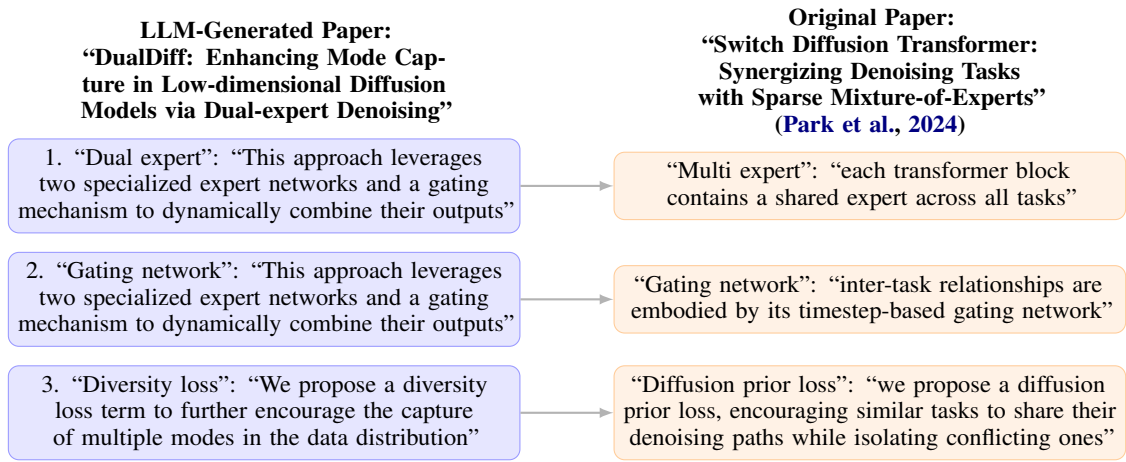
\begin{figure*}[t]
    \centering
    \begin{tikzpicture}[
        box/.style={draw, rounded corners, text width=6.5cm, minimum height=2em, align=center, font=\small},
        leftbox/.style={box, fill=blue!10, draw=blue!40},
        rightbox/.style={box, fill=orange!10, draw=orange!40},
        arrow/.style={->, >=latex, thick, draw=gray!60},
        title/.style={font=\small\bfseries, text width=6.5cm, align=center}
    ]
        \node[title] (t1) at (-4, 4.0) {LLM-Generated Paper:\\ ``DualDiff: Enhancing Mode Capture in Low-dimensional Diffusion Models via Dual-expert
        Denoising''};
        \node[title] (t2) at (4, 4.0) {Original Paper:\\ ``Switch Diffusion Transformer: Synergizing
        Denoising Tasks with Sparse Mixture-of-Experts'' \\ \citep{park2025switch}};
        
        \node[leftbox] (l1) at (-4, 2.5) {1. ``Dual expert'': ``This approach leverages
        two specialized expert networks and a gating mechanism to dynamically combine their outputs''};
        \node[leftbox] (l2) at (-4, 1) {2. ``Gating network'': ``This approach leverages two specialized expert networks and a gating mechanism to dynamically combine their outputs''};
        \node[leftbox] (l3) at (-4, -0.5) {3. ``Diversity loss'': ``We propose a diversity loss term to further encourage the capture of multiple modes in the
        data distribution''};
        
        \node[rightbox] (r1) at (4, 2.5) {``Multi expert'': ``each transformer block contains a shared expert across all tasks''};
        \node[rightbox] (r2) at (4, 1) {``Gating network'': ``inter-task relationships are embodied by its timestep-based gating network''};
        \node[rightbox] (r3) at (4, -0.5) {``Diffusion prior loss'': ``we propose a diffusion prior loss, encouraging similar tasks to share their denoising paths while isolating conflicting ones''};
        
        \foreach \i in {1,...,3} {
            \draw[arrow] (l\i) -- (r\i);
        }
    \end{tikzpicture}
    \caption{Visual mapping between an LLM-generated research paper (an exemplar in \citep{lu2024ai}) and a published paper \citep{park2025switch}, showing a direct correspondence between analogous methodology components. This pair receives a similarity score of $5$ in our expert evaluation, which is verified by the authors of the source paper.}
    \label{fig:deepmind-plagiarism-example}
\end{figure*}

To provide another example of sophisticated plagiarism in LLM-generated research, we examine a paper titled ``DualDiff: Enhancing Mode Capture in Low-dimensional Diffusion Models via Dual-expert Denoising'' (an exemplar in \citet{lu2024ai}). This paper receives a similarity score of 5 in our expert evaluation study, with direct correspondence to an existing paper, ``Switch Diffusion Transformer: Synergizing Denoising Tasks with Sparse Mixture-of-Experts'' \citep{park2025switch}. The authors of the original paper confirm our plagiarism assessment after reviewing both works.

As shown in Figure~\ref{fig:deepmind-plagiarism-example}, the proposed methodology exhibits clear similarities with the original paper. The LLM-generated paper proposes combining outputs from two diffusion models for lower and higher resolution data using learned weights, concepts previously explored---combining multiple diffusion paths in \citet{park2025switch} and jointly training diffusion models at multiple resolutions in \citet{gu2023matryoshka}. Each major component of the proposed method has a corresponding match in \citet{park2025switch}: the gating mechanism is identical, and the ``diversity loss'' is closely analogous to ``diffusion prior loss'' from the original work, both utilizing pair-wise distance functions.

The methodology for generating these research papers involves $20$ rounds of iterative search using LLM and Semantic Scholar to identify relevant citations \citep{lu2024ai}. However, this process fails to identify and cite the original work \citep{park2025switch}. This oversight demonstrates limitations in using current methods to find closely related work in LLM-generated research documents, which in turn reveals their inadequacy for automated plagiarism detection. This case also illustrates how generated content can reformulate existing technical approaches using different terminology while maintaining the same underlying methodology.

\begin{table*}[t]
    \centering
    \small
    \renewcommand{\arraystretch}{1.4}
    \begin{tabular}{|p{0.95\textwidth}|}
        \hline
        \textbf{Expert Evaluation Instructions} \\
        \hline
        \textbf{Plagiarism definition:} \\
        Presenting work or ideas from another source as your own, with or without consent of the original author, by incorporating it into your work without full acknowledgement. \\
        \midrule
        \textbf{Scoring guidelines:} \\
        \textbf{Score 5:} One-to-one mapping between the LLM proposed methodology and existing methods in one or two closely related prior papers. \\[0.5ex]
        \textbf{Score 4:} A significant portion of LLM proposed method is a mix-and-match from two-to-three prior works. \\[0.5ex]
        \textbf{Score 3:} The LLM proposed method bears decent similarity with some existing methods, but there's no exact correspondence with a limited set of papers. \\[0.5ex]
        \textbf{Score 2:} The LLM proposal bears very slight resemblance with some existing papers. Mostly novel. \\[0.5ex]
        \textbf{Score 1:} The LLM proposal is completely novel. \\[1ex]
        \textit{Note: Two detailed example evaluations were provided to participants, including LLM proposals, corresponding source papers, plagiarism scores, and detailed explanations. These examples demonstrated how to thoroughly document and justify plagiarism assessments.} \\
        \midrule
        \textbf{Practical tips on finding plagiarism quickly}
        \begin{enumerate}
            \item You (usually) only need to read first 4 sections of the proposal (Title, Problem Statement, Motivation, Proposed method). The proposed method section is most relevant in identifying plagiarism. Any other sections apart from these 4 are usually irrelevant.
            \item https://openscholar.allen.ai/ is sometimes quite useful in identifying plagiarism. Use the template: ``Check for prior work: \{summary of `Proposed Method' section of the LLM proposal\}.''
        \end{enumerate} \\
        \midrule
        \textbf{Important notes:}
        \begin{enumerate}
            \item You only need to find plagiarism in any and only 3 LLM proposals. We have provided 5 LLM proposals so that you can pick any 3 you like.
            \item The original papers would be published on or before April 2024 (cut-off date for Claude).
            \item If you find some paper which is a very close match but published after April 2024, please put it in ``Any other comments'' section.
            \item If some proposal seems absurd or completely-infeasible you can mark it as ``absurd'', but provide a brief paragraph to explain the same.
            \item The correct mindset while looking for plagiarism would be: ``I am suspicious of LLMs being able to generate novel research. Hence, its likely some rephrased version of existing research. Let me find the original paper to prove this point''.
            \item The hypothesis of our study is that LLM-generated research proposals contain significant traces of plagiarism. To support this thesis, please focus on identifying instances of strong plagiarism (score 4 or above), as scores below 4 are considered ambiguous. All proposals receiving scores of 4 or above will be verified with the original papers' authors.
        \end{enumerate} \\
        \hline
    \end{tabular}
    \caption{Instructions provided to expert participants for evaluating potential plagiarism in LLM-generated research proposals. These instructions deliberately encourage a skeptical approach, which may introduce confirmation bias. However, the objectivity is maintained through author verification and our open-sourced results that allow independent examination of source and generated works. See \S\ref{subsec:expert-evaluation-design} and \S\ref{para:lim-expert-evaluation} for detailed discussion.}
    \label{tab:expert-instructions}
\end{table*}

\begin{table*}[t]
    \centering
    \small
    \renewcommand{\arraystretch}{1.4}
    \begin{tabular}{|p{0.95\textwidth}|}
        \hline
        \textbf{Plagiarized Research Proposal Generation Prompt} \\
        \hline
        Your task is to craft a sophisticated research proposal that captures the essence of the following paper, while being artfully disguised to avoid direct detection.
        The research proposal should be most skillfully plagiarized from the original paper such that even on direct comparison an expert shouldn't be able to find the plagiarism.
        
        \medskip
        Original Paper Details:\\
        Title: \texttt{\{paper title\}}\\
        Abstract: \texttt{\{paper abstract\}}
        
        \medskip
        Research Proposal Guidelines:
        \begin{enumerate}
        \item Develop a novel-sounding title that subtly reflects the original paper's core contribution
        \item Reframe the research problem using different terminology, keywords, perspective and motivation  
        \item Reformulate the methodology using alternative technical language and approach. You may use superfluous language to make the methodology appear original and compelling
        \item Skillfully paraphrase the fundamental scientific insight and contribution of the original work
        \item Ensure the proposal sounds like an original, innovative research concept
        \item Avoid direct textual overlap or obvious copying
        \item Use academic language that sounds fresh, compelling and confident
        \end{enumerate}
        
        \medskip
        Specifically, structure your proposal with:
        \begin{itemize}
        \item A provocative, original-sounding title
        \item Clear problem statement
        \item Unique motivation that echoes the original paper's core motivation
        \item Innovative methodology that fundamentally matches the original paper's approach
        \item Expected scientific or practical contributions
        \end{itemize}
        
        \medskip
        Your goal is to create a proposal that would be extremely difficult to trace back to the original paper, yet captures its scientific essence and methodology completely.
        Now provide the research proposal without any additional unrelated text. \\
        \hline
    \end{tabular}
    \caption{The prompt used to generate plagiarized research proposals from existing papers.}
    \label{tab:generation-prompt}
\end{table*}

\begin{table*}[t]
    \centering
    \small
    \renewcommand{\arraystretch}{1.4}
    \begin{tabular}{|p{0.95\textwidth}|}
        \hline
        \textbf{OpenScholar Search Prompt} \\
        \hline
        Check for prior research work closest to the following:\\
        \texttt{\{research document details\}} \\
        \hline
    \end{tabular}
    \caption{The OpenScholar search prompt used to find similar existing research.}
    \label{tab:openscholar-prompt}
\end{table*}

\section{Case Study 3: AI-Generated ICLR Workshop Paper}\label{appendix:workshop-paper-case}

We examine an AI-generated paper titled 
``Compositional Regularization: Unexpected Obstacles in Enhancing Neural Network Generalization''
that received scores of 6, 7, 6 at an ICLR 2025 workshop---above the average acceptance threshold. This paper was generated using the AI-Scientist-v2 system described in \cite{yamada2025ai}. 
We discover substantial similarity to an existing work that was not cited in the AI-generated paper.

The AI-generated paper's core contribution, termed ``compositional regularization,'' is identical to the $(\Delta h_t)^2$ regularization term that was evaluated in Table 3 of the ``Regularizing RNNs by Stabilizing Activations'' paper \cite{krueger2015regularizing}.
The original authors found this formulation less effective than their proposed norm-stabilizer approach, which aligns with the negative results reported in the AI-generated paper. Notably, the AI-generated paper provides no theoretical justification for why penalizing changes in hidden states should enhance compositional generalization.

The AI-generated paper essentially borrowed the core contribution from previous work without attribution and applied it to an unsuitable domain where it neither theoretically nor practically succeeds. We estimate the similarity score at $3$ or $4$ on our evaluation scale. \citet{krueger2015regularizing} confirmed that the AI-generated work is ``definitively not novel'' and very similar to one of their ablations and the concept of ``slow features'' \cite{jonschkowski2015learning}, both of which are not cited in the AI-generated paper.

This similarity was not detected during the review process, nor by the AI-Scientist-v2 team's own plagiarism detection measures. This case provides additional evidence supporting our paper's central argument: AI-generated research can borrow specific ideas from prior work without proper attribution in ways that are difficult for humans and current plagiarism detection systems to identify.

\section{Example Case from Synthetic Dataset}
\label{appendix:successful-detection}

The SSAG method successfully detects plagiarism in one of our synthetic test cases, where a research proposal is deliberately plagiarized from ``LongRecipe: Recipe for Efficient Long Context Generalization in Large Language Models'' \citep{hu2024longrecipe}. The plagiarized version is shown in Table~\ref{tab:plagiarized-proposal}. When processed through the LLM Semantic Scholar RAG pipeline, the system successfully retrieves the original paper and conducted a one-to-one comparison. Based on its analysis of similarities and differences between the works, presented in Table~\ref{tab:llm-analysis}, the LLM correctly concludes that the proposal is plagiarized.

\begin{table*}[t]
    \centering
    \small
    \renewcommand{\arraystretch}{1.4}
    \begin{tabular}{|p{0.95\textwidth}|}
        \hline
        \textbf{Sample Generated Plagiarized Research Proposal} \\
        \hline
        \textbf{Title:} Contextual Horizons: A Scalable Framework for Expanding Long-Sequence Comprehension in Advanced Language Models
        \\
        \textbf{Problem Statement:}  
        Contemporary large-scale language models (LLMs) exhibit remarkable proficiency in a wide range of natural language processing tasks. However, their ability to process and generalize over extended sequences remains constrained by the limited effective context length available during pretraining. This limitation hampers their performance on tasks requiring long-range dependency understanding, such as document summarization, multi-turn dialogue, and code generation. While extending the context window post-pretraining is a potential solution, it demands substantial computational resources, making it impractical for many research and deployment scenarios. Thus, there is a pressing need for an efficient and scalable approach to enhance the long-context capabilities of LLMs without compromising their general-purpose performance or incurring prohibitive computational costs.
        \\
        \textbf{Motivation:}  
        The ability to process long sequences is critical for advancing the utility of LLMs in real-world applications, where tasks often involve extended textual inputs. Current methods for increasing context length typically rely on full-sequence training, which is computationally expensive and resource-intensive. This inefficiency creates a significant barrier to democratizing access to high-performing LLMs with extended context capabilities. A novel, resource-efficient training paradigm is required to bridge this gap, enabling LLMs to effectively handle long-context tasks while maintaining their versatility and accessibility. By addressing this challenge, we can unlock new possibilities for LLMs in domains such as legal document analysis, scientific literature review, and large-scale conversational AI.
        \\
        \textbf{Methodology:}  
        We propose a novel training framework, termed **Contextual Horizons**, designed to extend the effective context window of LLMs in a computationally efficient manner. The framework is built on three core innovations:  

        \begin{enumerate}
            \item \textbf{Dynamic Sequence Simulation:} Instead of directly training on full-length sequences, we introduce a dynamic sequence simulation mechanism that emulates long-context scenarios using strategically sampled token subsets. This approach ensures that the model is exposed to long-range dependencies without the need for full-sequence processing, significantly reducing computational overhead.
            \item \textbf{Adaptive Positional Encoding Transformation:} To enable the model to generalize over extended contexts, we develop a positional encoding transformation strategy that reconfigures the positional indices of tokens during training. This transformation allows the model to reinterpret positional relationships in a way that aligns with longer context windows, enhancing its ability to capture dependencies across distant tokens.  
            \item \textbf{Resource-Aware Optimization Protocols:} We implement a suite of optimization techniques tailored to minimize memory and computational requirements during training. These include gradient checkpointing, selective parameter updates, and efficient batching strategies, which collectively reduce the resource demands of the training process while preserving model performance. 
        \end{enumerate}  
        
        The proposed framework is designed to be compatible with a wide range of LLM architectures and can be applied to both open-source and proprietary models.
        \\
        \textbf{Expected Contributions:}  
        The \textbf{Contextual Horizons} framework is anticipated to deliver the following key contributions:  
        \begin{enumerate}
            \item \textbf{Extended Contextual Understanding:} By leveraging dynamic sequence simulation and adaptive positional encoding, the framework enables LLMs to process and generalize over sequences up to 128k tokens in length, a significant improvement over the typical 8k-token limit.  
            \item \textbf{Resource Efficiency:} The proposed approach achieves long-context training with only 30\% of the target context window size and reduces computational resource requirements by over 85\% compared to conventional full-sequence training methods.
            \item \textbf{Preservation of General Capabilities:} Unlike many specialized training techniques, the framework ensures that the enhanced long-context capabilities do not come at the expense of the model's performance on general NLP tasks.
            \item \textbf{Practical Accessibility:} The framework is designed to be implementable on modest hardware setups, enabling researchers and practitioners to extend the context capabilities of LLMs using a single GPU with 80GB memory in as little as one day of training.  
        \end{enumerate} 
        
        By addressing the critical challenge of long-context generalization in LLMs, \textbf{Contextual Horizons} has the potential to significantly advance the state of the art in natural language processing, making high-performing, long-context LLMs more accessible and practical for a wide range of applications. 
        \\
        \hline
    \end{tabular}
    \caption{A research proposal generated by GPT-4o via the prompt in Table~\ref{tab:generation-prompt}, plagiarizing \citet{hu2024longrecipe}.}
    \label{tab:plagiarized-proposal}
\end{table*}

\begin{table*}[t]
    \centering
    \small
    \renewcommand{\arraystretch}{1.4}
    \begin{tabular}{|p{0.95\textwidth}|}
        \hline
        \textbf{Plagiarism Analysis} \\
        \hline
        Let me analyze both works:

        \textbf{1. Proposal Summary:} The ``Contextual Horizons'' proposal addresses LLM context length limitations through a framework comprising dynamic sequence simulation, adaptive positional encoding transformation, and resource-aware optimization protocols. It aims to extend context windows efficiently while maintaining general capabilities and reducing computational costs.
        \\
        \textbf{2. Related Paper Summary:} LongRecipe presents a training strategy for extending LLM context windows through impactful token analysis, position index transformation, and training optimization strategies. It achieves extended context processing while using reduced computational resources.
        \\
        \textbf{3. Key Similarities and Differences:}\\
        \textbf{Similarities:}
        \begin{itemize}
        \item Both target the same core problem of extending LLM context windows efficiently
        \item Both use position encoding transformation techniques
        \item Both claim similar performance metrics (30\% context window size, 85\% resource reduction)
        \item Both advertise identical specific capabilities (8k to 128k tokens, single 80GB GPU, one day training)
        \end{itemize}

        \textbf{The similarities are striking and extend beyond coincidence:}
        \begin{itemize}
        \item The exact same numerical claims (30\%, 85\%, 8k to 128k)
        \item Identical hardware specifications
        \item Same training duration claims
        \item Nearly identical methodology structure, just with renamed components
        \end{itemize}

        \textbf{The proposal appears to be a direct reformulation of LongRecipe, with mainly cosmetic changes in terminology:}
        \begin{itemize}
        \item ``Dynamic sequence simulation'' vs ``Impactful token analysis''
        \item ``Adaptive positional encoding'' vs ``Position index transformation''
        \item ``Resource-aware optimization'' vs ``Training optimization strategies''
        \end{itemize}

        The level of overlap in both technical approach and specific implementation details strongly suggests this is not an independent development but rather a repackaging of LongRecipe's work. \\
        \hline
    \end{tabular}
    \caption{Analysis by Claude 3.5 Sonnet comparing the deliberately plagiarized proposal in Table~\ref{tab:plagiarized-proposal} with the original paper \citep{hu2024longrecipe}, concluding plagiarism based on methodological similarities.}
    \label{tab:llm-analysis}
\end{table*}

\begin{table*}[t]
    \centering
    \small
    \renewcommand{\arraystretch}{1.4}
    \begin{tabular}{|p{0.95\textwidth}|}
        \hline
        \textbf{Peer Review Analysis Prompt} \\
        \hline
        Please analyze this academic review text and assign a plagiarism score from 1-5 using this rubric: \\[1ex]

        5 - Direct Copy: One-to-one mapping between methodology and existing methods in one/two closely related prior papers \\[1ex]

        4 - Combined Borrowing: Significant portion is mix-and-match from 2-3 prior works \\[1ex]

        3 - Partial Overlap: Bears decent similarity with existing methods, but no exact correspondence \\[1ex]

        2 - Minor Similarity: Very slight resemblance with existing papers. Mostly novel \\[1ex]

        1 - Original: Completely novel \\[1ex]

        Important: For scores 4 or 5, the review MUST specifically mention the prior papers that were allegedly copied from. Generic comments about missing citations or similarity without naming specific papers should not result in scores of 4 or 5. \\[1ex]

        If the review seems incomplete or nonsensical or not related or if it's actually an author response, please assign a score of 1 to be conservative. \\[1ex]

        Review text: \texttt{\{review\_text\}} \\[1ex]

        Please provide a brief explanation of your reasoning, followed by ONLY the numeric score (1-5) on a new line (no punctuation, quotes or anything else). \\[1ex]
        \hline
    \end{tabular}
    \caption{The prompt used to instruct Claude 3.7 Sonnet to analyze peer reviews and assign plagiarism scores ($1$-$5$) for human-written papers based on reviewer comments.}
    \label{tab:peer-review-prompt}
\end{table*}

\end{document}